\documentclass{article}

\usepackage[dblblindworkshop, final]{neurips_2025}

\usepackage[utf8]{inputenc} 
\usepackage[T1]{fontenc}    
\usepackage{hyperref}       
\usepackage{url}            
\usepackage{booktabs}       
\usepackage{amsfonts}       
\usepackage{amsmath}
\usepackage{amssymb}
\usepackage{nicefrac}       
\usepackage{microtype}      
\usepackage{xcolor}         

\usepackage{nameref}
\usepackage[shortlabels]{enumitem}
\usepackage{tcolorbox}
\usepackage{url}
\usepackage{multirow}
\usepackage[noabbrev,nameinlink]{cleveref}
\usepackage{bm}

\definecolor{sns_blue}{HTML}{4c72b0}
\definecolor{sns_orange}{HTML}{dd8452}
\definecolor{sns_green}{HTML}{55a868}
\definecolor{sns_red}{HTML}{c44e52}

\title{Towards Leveraging Sequential Structure\\in Animal Vocalizations}
\workshoptitle{AI for Non-Human Animal Communication}

\author{Eklavya Sarkar\\
Idiap Research Institute\\
Ecole Polytechnique Fédérale de Lausanne\\
Switzerland \\
\texttt{eklavya.sarkar@idiap.ch} \\
\And
Mathew Magimai.-Doss, \\
Idiap Research Institute \\
Switzerland \\
\texttt{mathew@idiap.ch} \\
}

\begin{document}

\maketitle

\begin{abstract}
  Animal vocalizations contain sequential structures that carry important communicative information, yet most computational bioacoustics studies average the extracted frame-level features across the temporal axis, discarding the order of the sub-units within a vocalization. This paper investigates whether discrete acoustic token sequences, derived through vector quantization and gumbel-softmax vector quantization of extracted self-supervised speech model representations can effectively capture and leverage temporal information. To that end, pairwise distance analysis of token sequences generated from HuBERT embeddings shows that they can discriminate call-types and callers across four bioacoustics datasets. Sequence classification experiments using $k$-Nearest Neighbour with Levenshtein distance show that the vector-quantized token sequences yield reasonable call-type and caller classification performances, and hold promise as alternative feature representations towards leveraging sequential information in animal vocalizations.
  
\end{abstract}

\section{Introduction}
The communicative power of sequences in animal vocalizations is well-documented across species, with vocal sequences often serving key biological roles such as territory defense, mate attraction, social bonding, and alarm signaling \citep{kershenbaum2014acoustic}. The complexity of these sequences manifests through distinct patterns of acoustic units that are combined in species-specific ways, following implicit or explicit syntactic rules. For instance, songbirds produce vocalizations composed of repeated motifs and notes arranged in recognizable patterns \citep{catchpole2003birdsong}, while cetaceans exhibit intricate, temporally-structured acoustic sequences associated with social interaction and individual identification \citep{mercado2012structure}. Thus, capturing and analyzing the inherent sequential structure in animal vocalizations could substantially enhance our understanding of their communicative function and biological significance.

However, in many existing computational bioacoustics works \citep{sarkar23_interspeech, sarkar24_vihar, mahoud24_vihar, Sarkar_Bioacoustics_Journal, Sarkar_Idiap-RR-05-2025}, each sample's extracted feature embeddings $\bm{x}\in\mathbb{R}^{N\times D}$ are typically averaged into a vocalization-level representation, denoted as a functional vector $f_{\mu} = \mu(\bm{x}) \in \mathbb{R}^{D}$ or $\bm{f}_{\mu\sigma} = \bigl[\mu(\bm{x}),\,\sigma(\bm{x})\bigr]\in\mathbb{R}^{2D}$. While these `stats‐pooled' representations have proven very valuable for classification tasks, bandwidth analysis, and model adaptation, they ignore the sequential aspect of animal calls: each vocalization is treated like an unordered bag of frame-level feature embeddings. This completely overlooks the fact that many animal arrange acoustically distinct sub-vocalization units in a specifically ordered sequences that carry important communicative and syntactic information \citep{AcousticSequencesInAnimals}. The goal of this paper is thus to investigate alternate feature representations that can capture the sequential structure within animal vocalizations, and leverage the unutilized temporal information to improve classification performance. 

In order to effectively model sub-vocalization unit level sounds, we turn to symbolic speech tokenization. Recent work has shown that discrete audio tokens obtained through vector-quantization of `continuous' SSL feature embeddings can effectively encode acoustic information, and thus be utilized for many speech and audio tasks \citep{guo2025_tokens}. Based on this prior, we extend this framework to bioacoustics, and explore whether discretization of animal vocalizations into token sequences can reveal meaningful structure and help distinguish call-types or individual callers. A successful framework could even yield an inventory of recurring acoustic sub‐vocalization units in animal communication. To the best of our knowledge, this is the first work to explore discrete audio tokens for bioacoustics. To that end, we investigate different discretization methods, namely vector quantization and gumbel-softmax vector quantization, through a distance analysis, and then benchmark the downstream classification performance to evaluate their efficiency towards leveraging the sequential structure in animal calls.


The rest of this paper is organized as follows: \Cref{sec:tokens} presents a review of representation learning using discrete audio tokens, \Cref{sec:exp_setup} describes our experimental setup, and \Cref{sec:results} presents the results and analysis. Finally, \Cref{sec:tokens_conclusion} concludes the paper with directions for future work.

\section{Discrete Audio Tokens-based Representation Learning}
\label{sec:tokens}

Modern self-supervised learning (SSL) foundation models pre-trained on human speech have shown strong transferability to bioacoustic tasks \citep{sarkar_thesis}. Many such SSL models employ discrete token representations during their pre-training stages which are typically derived using a discretization process, either through integrated vector quantization layers \citep{Baevski2020vq-wav2vec, baevski2020wav2vec2} or offline clustering mechanisms applied to continuous embeddings \citep{hsu2021hubert}. However, such discrete representations are primarily intended to facilitate SSL objectives, and are usually not directly exposed or utilized during inference or downstream tasks. In this paper, we explicitly leverage the discrete tokenization methodology. To do so, we first extract window-level embeddings from a pre-trained SSL model, and subsequently train a separate quantization module which maps the embeddings into sequences of discrete tokens. Note that the quantization is performed independently \textit{per frame}, thereby preserving the temporal order of the original acoustic events within the vocalization, and is trained separately on extracted embeddings from the pre-trained encoder, using the bioacoustic data of interest. This would allow the codebook vectors to adapt specifically to the acoustic characteristics and distributions of the calls being studied.

\begin{figure}[htb]
  \centering
  \includegraphics[width=\linewidth]{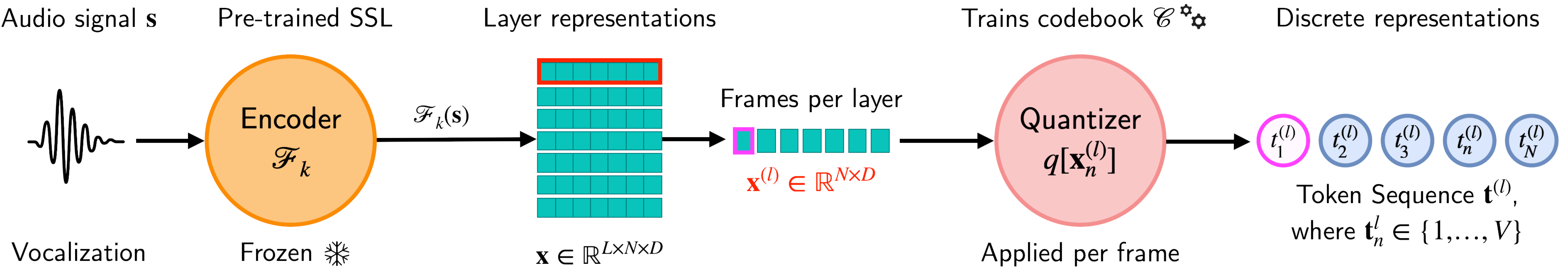}
  \caption{Discrete call tokenization pipeline using vector quantization.}
  \label{fig:tokens_pipeline}
\end{figure}

The overall call tokenization pipeline, employed in this work, using vector quantization is illustrated in \Cref{fig:tokens_pipeline}. Specifically, a raw audio waveform $\bm{s}$ is first passed through a pre-trained encoder $\mathcal{F}$, producing continuous layer embeddings $\bm{x} \in \mathbb{R}^{L \times N \times D}$, where $L$ is the number of layers, $N$ the number of frames in each layer, and $D$ the dimension of each frame. Let $\bm{x}_n^{(l)} \in \mathbb{R}^D$ denote the embedding extracted from encoder layer $l$ at frame position $n$. Each layer embedding is then quantized individually per-frame by a quantization function $q$, resulting in discrete tokens $\bm{t}^{(l)}_n = q[\bm{x}_n^{(l)}]$. Formally, the quantization function maps each embedding from continuous $D$-dimensional space to a discrete integer token index $q: \mathbb{R}^D \rightarrow \{1, 2, \dots, V\}$ where $V$ denotes the vocabulary size, i.e., the number of unique discrete tokens. Each token index corresponds directly to an entry in a finite set, referred to as the codebook $\mathcal{C} = \{\bm{c}_1, \bm{c}_2, \dots, \bm{c}_V\}$, where each code-vector $\bm{c}_i \in \mathbb{R}^D$ corresponds to the $i$-th discrete token in the original embedding space.

\section{Experimental Setup}
\label{sec:exp_setup}
Common marmosets (\textit{Callithrix jacchus}) are highly vocal new world primates with an extensive vocal repertoire, often studied to understand the evolutionary origins of human vocal communication. For our study, we conducted our experiments on the three marmosets datasets from \citep{Sarkar_Bioacoustics_Journal}, namely the \textit{InfantMarmosetsVox (IMV)} \citep{sarkar23_interspeech}, \textit{Bosshard} \citep{Alex_MA_thesis, bosshard_2024}, and \textit{Wierucka} datasets. We also included the dogs dataset \citep{abzaliev24}, referred to as \textit{Abzaliev}. For further dataset details, the reader is referred to \Cref{sec:datasets}. To evaluate performance, we look at the call-type (CTID) and caller (CLID) classification tasks. Given the demonstrated transferability of learnt speech representations to animal vocalizations \citep{sarkar_thesis}, we select HuBERT as our pre-trained SSL model based on its existing benchmarking and effectiveness on bioacoustics tasks \citep{Sarkar_ICASSP_2025}, and first extract embeddings $\bm{x} \in \mathbb{R}^{L \times N \times D}$ from the raw input calls.

To discretize the animals calls we investigate vector quantization (VQ) and gumbel-softmax vector quantization (GVQ) due to their proven effectiveness in quantizing audio embeddings. Their respective implementations are detailed in \Cref{ssec:vq} and \ref{ssec:gvq}. We first train them on $\bm{x}$ based on our established protocols (\ref{ssec:tokens_train_protocol}), and then proceed to generate and post-process token sequences $\bm{t}$ for all data samples (\ref{ssec:tokens_seq_gen}). Note that most modern acoustic tokenizers have multiple quantizers, however, for simplicity and clarity, we focus only on hand-coded single-codebook ones in this work.


%

As our first investigation step, we examine the intra and inter-class distances between all the generated token sequences to assess their degree of separability. We compute all the pairwise Levenshtein distances, categorizing comparisons into: (i\textcolor{sns_orange}{$\bullet$}) \textit{Intra‑caller, intra‑calltype}: two vocalizations from the same caller producing the same call-type (smallest expected distance). (ii\textcolor{sns_blue}{$\bullet$}) \textit{Intra‑caller, inter‑calltype}: two vocalizations from the same caller producing different call-types. (iii\textcolor{sns_green}{$\bullet$}) \textit{Inter‑caller, intra‑calltype}: two vocalizations from different callers producing the same call-type. (iv\textcolor{sns_red}{$\bullet$}) \textit{Inter-caller, inter-calltype}: two vocalizations from different callers producing different call-types (largest expected distance).


Based on the insights of the comparative analysis, we then evaluate how well the sequential nature of token representations can be leveraged for CTID and CLID. We classify the token sequences with the $k$-NN algorithm, using the Levenshtein distance as the similarity metric. The predicted label of a sample is determined by applying a majority-voting framework on the actual labels of the $k$ most similar sequences. The classifier training and hyperparameter details are given in \ref{sec:classif_exp_setup}. Finally, we evaluate this tokenization approach to a `traditional' linear-probing baseline, i.e. by pooling the temporal information into a functional $\bm{f}_{\sigma\mu} \in \mathbb{R}^{2D}$, and classifying it with a fully-connected layer.

\section{Results}
\label{sec:results}

\subsection{Distance Analysis}
\label{sec:tokens_compar_analysis}

\begin{figure}[h]
  \centering
  \includegraphics[width=\linewidth]{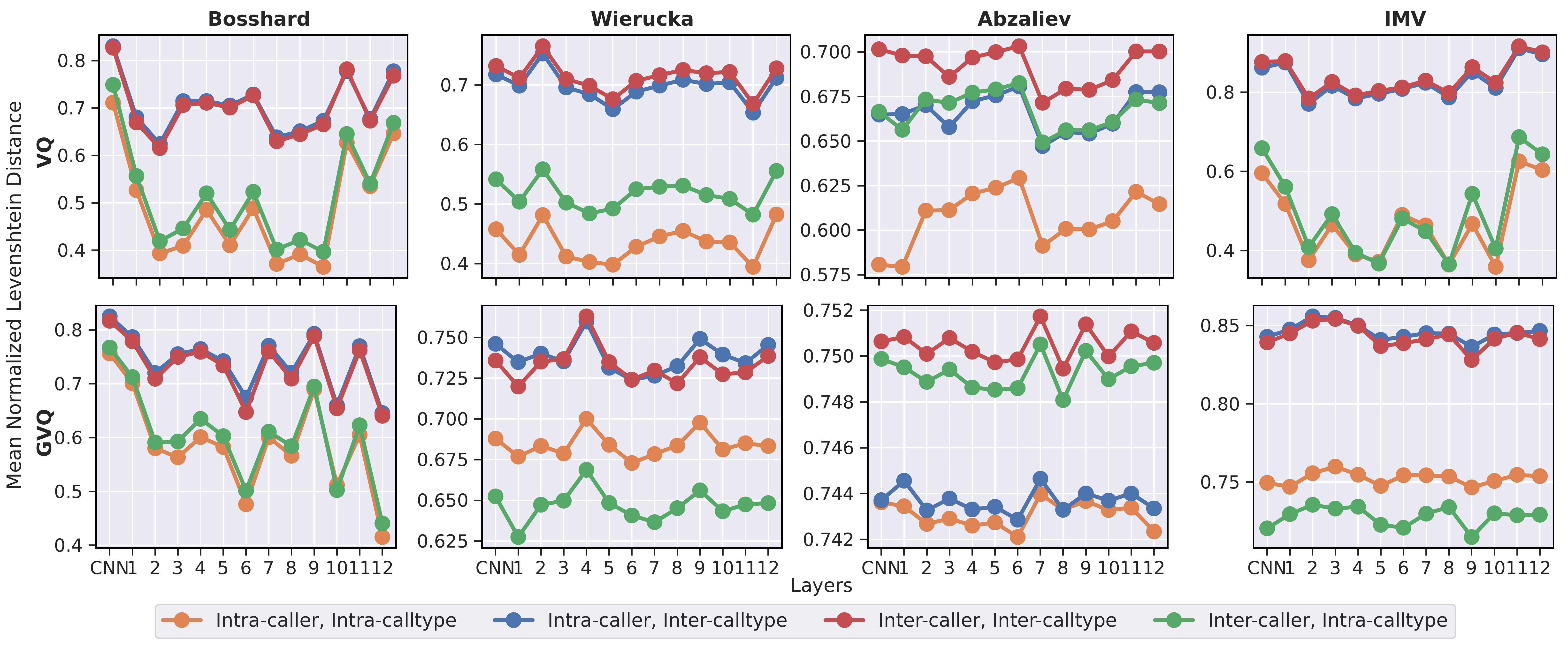}
  \caption{Layer-wise mean Levenshtein distance between all pairs of VQ and GVQ token sequences.}
  \label{fig:tokens_edit_distances_vq}
\end{figure}

\Cref{fig:tokens_edit_distances_vq} presents the means of the distances distributions of the four aforementioned categories. For the token sequences generated from the VQ model (top row) we can observe that groups (i \textcolor{sns_orange}{$\bullet$}) and (iv \textcolor{sns_red}{$\bullet$}) behave as expected: they both have the smallest and largest distance, on average, for all datasets. We also noticeably observe that group (ii \textcolor{sns_blue}{$\bullet$})'s distance is larger than group (iii \textcolor{sns_green}{$\bullet$})'s for most datasets. This makes sense intuitively: two vocalizations produced by the a caller vocalizing different call-types are more likely to be acoustically distinct, than two generated by different callers vocalizing the same call-type. The discrete acoustic tokens sequences reflect this distribution, demonstrating their ability to model and capture the temporal information encoded in vocalizations.


While we can observe similar trends with the GVQ tokens (bottom row) on Bosshard, the remaining datasets exhibit some different patterns. Notably, group (ii \textcolor{sns_blue}{$\bullet$}) and (iii \textcolor{sns_green}{$\bullet$})'s trends are flipped in Abzaliev. This may be due to the comparatively large number of callers (80), which increases acoustic variability and makes it harder to distinguish sequences of the same call produced by different callers than those of different calls produced by the same caller. Additionally, for Wierucka and IMV, (iii \textcolor{sns_green}{$\bullet$}) is unexpectedly smaller on average than (i \textcolor{sns_orange}{$\bullet$}). This suggests that the GVQ tokens do not consistently preserve fine-grained caller-specific information as well as the VQ tokens across all datasets.

This analysis shows that VQ tokens are indeed capable of clustering sufficient acoustic information to discriminate calls or callers, under real-world left-to-right temporal constraints. The GVQ tokens exhibit some unexpected patterns and less consistent separability, indicating they may be less effective. 






\subsection{Classification Analysis}
\label{sec:tokens_classif}

\begin{figure}[ht]
  \centering
  \includegraphics[width=\linewidth]{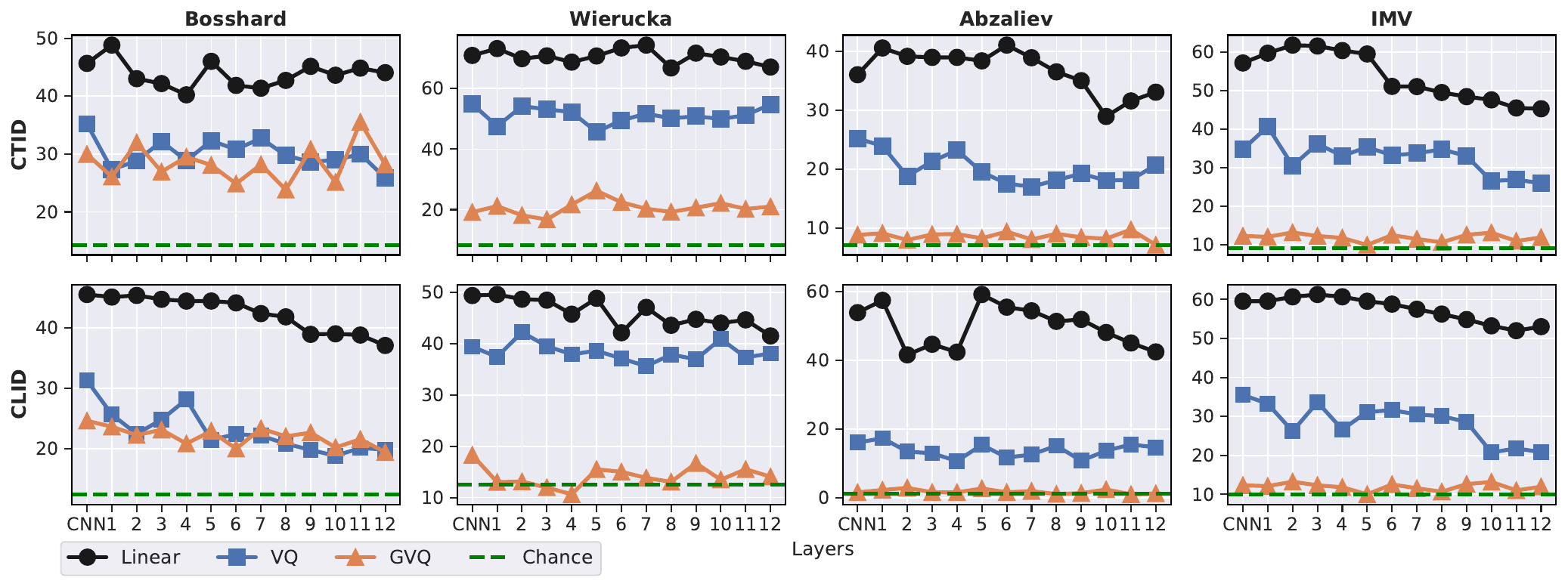}
  \caption{Layer-wise UAR [\%] for CTID (top) and CLID (bottom) using $k$-NN on token sequences.}
  \label{fig:tokens_knn_scores}
\end{figure}

The classification results are shown in \Cref{fig:tokens_knn_scores} using the VQ and GVQ tokens. We can observe that the linear baseline clearly outperforms the $k$-NN classification of token sequences in all scenarios, showing that the HuBERT feature embeddings currently capture more meaningful information. GVQ notably underperforms on all tasks and datasets, especially Abzaliev, essentially achieving chance-level performance, strongly suggesting a codebook collapse to a restricted subset of symbols. On the other hand, VQ yields a weaker performance than the linear baseline, but still achieves reasonable results in many scenarios. This indicates that some degree of call or caller-level discrimination can already be captured simply with the Levenshtein distance, as also seen in \Cref{sec:tokens_compar_analysis}, but substantial improvements in classification performance could potentially be achieved through more sophisticated sequence modeling of the generated token sequences. Finally, while a single shared codebook can still encode enough information for call discrimination, it is perhaps not expressive enough to preserve the finer caller‑specific nuances existing in the continuous embeddings.


\Cref{tab:comb_scores} tabulates the highest scores of each feature across layers, and also shows the drop in performance, denoted with $\Delta$, of the token sequence-based representations compared to the linear baseline. Similar to the results in previous chapters, we can see that the CTID classification yields higher scores than CLID across all feature representations. This highlights that call‑types differ in distinct spectro‑temporal patterns that token sequences can still capture, where as caller identity is largely carried by subtler characteristics that are harder to preserve after vector quantization. This also suggests that discrete tokens need a higher-resolution to be effective.

\begin{table}[ht]
    \centering
    \caption{Best UAR [\%] scores for each feature across layers. $n_C$ is the number of classes for that dataset and task, and chance performance is calculated as 100/$n_c$. $\bm{\Delta}$ represents the relative drop in performance with respect to the linear layer baseline.}
    \begin{tabular}{llrrrrrrr}
    \toprule
    \textbf{Task} & \textbf{Dataset} & $\bm{n_C}$ & \textbf{Chance} & \textbf{Linear} & \textbf{VQ} & \textbf{GVQ} & $\bm{\Delta}$\textbf{VQ} & $\bm{\Delta}$\textbf{GVQ} \\
    \midrule
    \multirow{4}{*}{CTID}
    & Bosshard &  7 & 14.30 & 48.81 & 35.20 & 35.52 & 27.88 & 27.23 \\
    & Wierucka & 12 &  8.30 & 74.36 & 54.91 & 26.23 & 26.16 & 64.72 \\
    & Abzaliev & 14 &  7.14 & 41.07 & 25.24 & 9.78 & 38.54 & 76.20 \\
    & IMV      & 11 &  9.10 & 61.75 & 40.65 & 24.94 & 34.17 & 59.60 \\
    \midrule
    \multirow{4}{*}{CLID}
    & Bosshard &  8 & 12.50 & 45.52 & 31.31 & 24.65 & 31.22 & 45.85 \\
    & Wierucka &  8 & 12.50 & 49.60 & 42.24 & 18.29 & 14.83 & 63.13 \\
    & Abzaliev & 80 & 1.25  & 59.09 & 17.35 & 2.90 & 70.64 & 95.09 \\
    & IMV      & 10 & 10.00 & 61.28 & 35.51 & 13.23 & 42.05 & 78.42 \\
    \bottomrule
    \end{tabular}
    \label{tab:comb_scores}
\end{table}

\begin{figure}[ht]
  \centering
  \includegraphics[width=\linewidth]{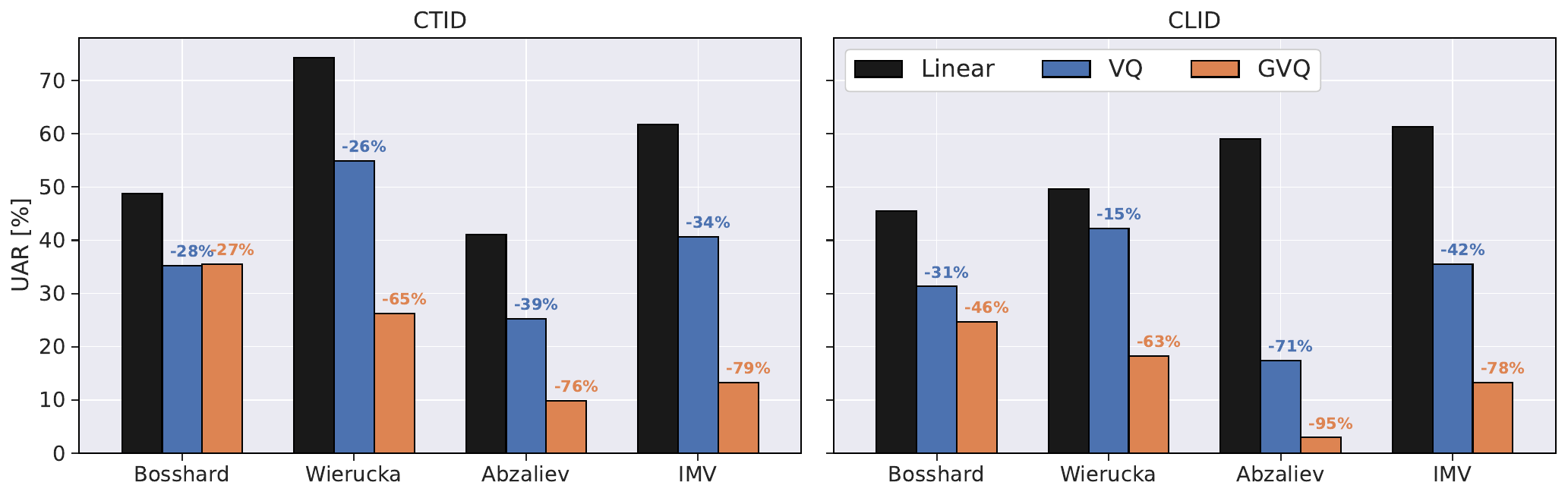}
  \caption{Best UAR results across layers for CTID and CLID.}
  \label{fig:comb_bar}
\end{figure}

\Cref{fig:comb_bar} visually plots the same information. For CTID, discretizing the feature embeddings with a VQ and GVQ drops the performance across datasets by $\sim$26-39\% and $\sim$27-79\% respectively, when compared to stats-pooling the same features and then classifying with a linear layer. For CLID, the drop is of $\sim$15-71\% and $\sim$46-95\% respectively. These strong decreases in performances reveal that perhaps a single VQ or GVQ codebook is not enough to effectively model the entire animal vocalizations alone, especially for CLID, or the arbitrary codebook size of $V=50$. In our early ablation experiments, however, we did not empirically observe a significant change in performance when compared to $V=25$ or $100$. A plausible next step could thus be to train a quantizer model which employs \emph{multiple} codebooks to retain a richer set of temporal patterns.

Lastly, although we trained a single codebook, shared across all layers, for both VQ and GVQ, we can still observe in \Cref{fig:tokens_knn_scores} that earlier layers tend to yield better performance across tasks, consistent with the trends reported in other works \citep{Sarkar_Bioacoustics_Journal, Sarkar_ICASSP_2025}. This indicates that differences between layers persist even after discretization, and that sharing a codebook does not diminish the higher capability of earlier layers in encoding salient and transferable representations.

\section{Conclusions and Future Work}
\label{sec:tokens_conclusion}
In this paper, we explored alternate feature representations that could preserve the temporal structure of animal vocalizations instead of averaging their extracted feature embeddings into single functional vectors. To that end, we investigated call discretization and evaluated whether discrete acoustic token modeling could effectively improve animal call classification performance. We first trained a vector quantization and a gumbel-softmax vector quantization module to convert vocalizations into discrete token sequences for four different animal datasets. A distance analysis of generated sequences showed that they are able to encode temporal information, and exhibit a degree of separability by call-type or caller identity across all datasets. Training a $k$-NN classifier on said representations showed that while VQ token representations are still weaker than linear-probing baselines, they are nonetheless able to leverage meaningful sequential information from animal vocalizations, and incorporating more sophisticated sequence modeling could further improve performance.

The scope for improvements on this topic is fairly substantial. One line of future work could explore larger, multi-codebook quantization architectures, such as Residual VQs \citep{1171604} or Grouped VQ \citep{5432202}. Another direction could investigate other sequence post-processing techniques, such as deduplication, i.e. removing consecutive duplicate tokens \citep{10447929}, or acoustic byte-pair encoding (BPE) \citep{bpe_algo}, which can further reduce the sequence length and tighten the alignment between tokens and acoustically meaningful sub-units.

\begin{ack}
This work was funded by the Swiss National Science Foundation’s NCCR Evolving Language project (grant no. 51NF40\_180888).
\end{ack}

\bibliographystyle{unsrt}
\bibliography{neurips_2025}

\appendix

\section{Quantization}

\subsection{Vector Quantization (VQ)}
\label{ssec:vq}

While traditional clustering methods operate independently of model training, vector quantization integrates a discrete, learnable codebook directly into the neural network \citep{oord2017neural}, enabling end-to-end optimization via gradient propagation through the quantization step.

We maintain a learnable codebook $\mathcal{C} = \{\bm{c}_1, \dots, \bm{c}_V\}\in \mathbb{R}^{V\times D}$ of $V=50$ code-vectors, each of dimension $D=768$. Given an input embedding $\bm{x}_n^{(l)}$, the quantization process selects the nearest codebook vector $\bm{c}_i$ by simply minimizing the Euclidean distance between the two:

\begin{equation}\label{eq:vq_dist}
q[\bm{x}_n^{(l)}] = \underset{i \in \{1, 2, \ldots, V\}}{\arg\min} \|\bm{x}_n^{(l)} - \bm{c}_i\|^2_2
\end{equation}

which returns the token index which is the input's discrete token. The codebook vector itself, which we denote as $\bm{c}_k \triangleq \bm{c}_{q(\bm{x})}$, is passed on to subsequent networks.

To allow backpropagation through the non‐differentiable nearest‐neighbor argmin lookup given in \ref{eq:vq_dist}, a \textit{straight‐through estimator} (STE) \citep{bengio2013ste} is employed to graft gradients from the quantized output \(\bm{c}_k\) back to \(\bm{x}_n^{(l)}\) during the backward pass. The encoder thus receives learning signals from downstream losses, while the codebook vectors themselves are updated via the VQ loss below. In our case, since we have pre-extracted embeddings, no encoder is updated, and the downstream losses encourage the extracted representations to align with their assigned code‐vectors, even though only the codebook parameters are updated. During training, we optimize the VQ loss which is jointly defined as the sum of the codebook and commitment losses:

\begin{equation}
\begin{aligned}
\mathcal{L}_{\text{VQ}} &= 
    \underbrace{\|\mathrm{sg}[\bm{x}_n^{(l)}] - \bm{c}_k\|_2^2}_{\text{Codebook Loss}}
  + \underbrace{\beta\,\|\bm{x}_n^{(l)} - \mathrm{sg}[\bm{c}_k]\|_2^2}_{\text{Commitment Loss}}.
\end{aligned}
\end{equation}

where \(\mathrm{sg}[\cdot]\) denotes the stop‐gradient operator and the beta coefficient is typically set to $\beta = 0.25$. The codebook loss shifts the selected code-vector \(\bm{c}_k\) toward its corresponding input embedding \(\bm{x}_n^{(l)}\), whereas the commitment loss conversely encourages the embedding to move closer to its matched codeword. We iterate $\mathcal{L}_{\mathrm{VQ}}$ over all the layers $L$ and frames $N$ to compute the total cost. While one can also update the codebook via an exponential-moving-average (EMA) scheme \citep{oord2017neural}, we focus here on the loss-based updates for clarity. Since the encoder is kept frozen, both terms in practice serve to adapt the codebook vectors to the distribution of the bioacoustic embeddings, yielding a discrete vocabulary that best captures their statistical structure.

VQs are unfortunately also known to suffer from codebook collapse, where the codebook usage is highly imbalanced, i.e. most input embeddings get mapped to a one or two centroids, while the rest of the codebook remains idle and unpdated, drastically reducing its effective representation capacity.

\subsection{Gumbel-Softmax Vector Quantization (GVQ)}
\label{ssec:gvq}

To mitigate codebook collapse in the standard VQ, we also implement gumbel vector quantization (GVQ) \citep{jang2017categorical}, which uses the gumbel–Softmax relaxation as a proxy for classic Softmax and to enable differentiable sampling from a categorical distribution. Given an input embedding $\bm{x}_n^{(l)}$, a linear projection layer computes logits $\{\pi_i\}_{i=1}^V$. The relaxed one‐hot vector $\bm{p}\in\Delta^{V-1}$ is then obtained via:

\begin{equation}
p_i = \frac{\exp\bigl((\log \pi_i + g_i)/\tau\bigr)}{\sum_{j=1}^V \exp\bigl((\log \pi_j + g_j)/\tau\bigr)}, 
\end{equation}

where each $g_i$ is an independent sample from the $\mathrm{Gumbel}(0,1)$ distribution and $\tau$ is a fixed temperature (set to 1.0).  A straight‐through estimator is applied so that, during the forward pass, the highest‐probability entry in $\bm{p}$ is discretized to a one‐hot vector, while in the backward pass gradients flow through $\bm{p}$ as if the operation were identity.

Training of the GVQ module is driven by an entropy‐maximizing loss that encourages uniform use of all $V$ codewords. Equivalently, this can be written as a KL divergence between $\bm{p}$ and the uniform distribution:
\begin{equation}
\mathcal{L}_{\mathrm{GVQ}}
= \sum_{i=1}^V p_i \log\bigl(p_i\,V\bigr)
\end{equation}

In our GVQ implementation, we implement several extensions to improve codebook utilization and robustness. First, we augment the KL divergence objective with a tunable weight parameter $\alpha_{\mathrm{KL}}$. Second, we add a diversity loss term weighted by a hyperparameter $\lambda_{\mathrm{div}}$, which explicitly penalizes under‐utilization of the codebook. Throughout training, we track two key metrics: the codebook perplexity
\begin{equation}
\mathrm{PPL} \;=\;\exp\!\Bigl(-\sum_{i=1}^V \bar p_i \log \bar p_i\Bigr),
\end{equation}
where $\bar p_i$ is the average probability of selecting codeword $i$, and the normalized perplexity $\mathrm{PPL}/V$. The diversity loss is defined to increase the normalized perplexity, thereby encouraging the model to make use of a larger fraction of available codewords.

\subsection{Quantizer Training Protocol}
\label{ssec:tokens_train_protocol}
We train all of our vector‐quantization models on $\bm{x}$ using the Adam optimizer with a fixed batch size of 32, running for up to 20 epochs on \textit{Train}, and evaluating performance on a held-out \textit{Val} set to monitor convergence and guard against overfitting. To find the best hyperparameter settings, we conduct a grid search over two quantizer variants, as given in \Cref{tab:vq_search_space}.


Note that for both quantizer models, the codebook $\mathcal{C}$ is \textit{shared} across all layers $L$ during training. Having the same symbol inventory for every layer makes the token sequences directly comparable across layers, and removes the need to have 13 separate vocabulary sets. Since the codebook must cover the union of all layer manifolds, a codebook-collapse is unlikely, and much less so than the alternate scenario of layer-specific sub-codebooks. 

Each mini‑batch therefore contains all layers of every utterance during training: batch tensors of shape ($B, L, N, D$), corresponding to the batch size, layer index, frame index, and feature dimension respectively, are reshaped to $(B \times L, N, D)$, quantized with a $V=50$ entry codebook, and then reshaped back. This allows the quantizer $q$ to see inputs from all layers, but then generate token sequences $\bm{t}$ drawn from the common symbol set.

\begin{table}[ht]
\centering
\caption{Hyperparameter search space for VQ and GVQ models.}
\begin{tabular}{lll}
\toprule
\textbf{Quantizer} & \textbf{Hyperparameter} & \textbf{Search Space} \\
\midrule
\multirow{3}{*}{VQ} & Learning rate & 1e[-4, -3, -2] \\
                    & Commitment cost & 0.25 \\
                    & EMA & [True, False] \\
\midrule
\multirow{7}{*}{GVQ} & Learning rate & 1e[-4, -3, -2] \\
                     & KL weight & [0.5, 1.0, 1.5, 2.0] \\
                     & Diversity weight & [0.0, 0.01, 0.05, 0.1, 0.2, 0.5] \\
                     & Temperature schedule: & \\
                     & \quad Max temperature & 2.0 \\
                     & \quad Min temperature & 0.1 \\
                     & \quad Decay factor & 0.999 \\
\bottomrule
\end{tabular}
\label{tab:vq_search_space}
\end{table}

\subsection{Token Sequence Generation and Post-Processing}
\label{ssec:tokens_seq_gen}
After training the quantizer on \textit{Train}, we generate and save sequences of acoustic discrete tokens $\bm{t}$ for each vocalization in the entire dataset as described in the pipeline in \Cref{sec:tokens}. However, during batch processing, audio waveforms are repeat-padded to match the length of the longest sample within the batch. This repetition artificially inflates all the token sequences except one to be longer than the actual audio signals. To account for this, we apply some post-processing to the sequence by first calculating the effective number of frames of each data sample. We determine the downsampling factor of a batch by dividing the longest raw audio length in a given batch by the number of frames in its token sequence. Then, for each data sample, we compute the effective frame count by dividing its raw audio length by this factor and rounding the result. Finally, the token sequence for each sample is trimmed to this effective frame count, yielding a variable-length representation that accurately reflects the original signal duration and excludes any tokens that result solely from the padding. To ensure consistency with the original embedding extraction process, we implement verification mechanisms that confirm sample ordering is maintained throughout the token generation pipeline.


\section{Datasets}
\label{sec:datasets}

\Cref{table:dataset_stats} presents a statistical summary of the datasets used in this work. Marmosets are a central focus of this paper, as their vocal behaviour provides a particularly valuable surrogate model for studying the evolutionary origins of human language. Indeed, their relevance to comparative communication science makes them especially well-suited for exploring how vocal signals encode socially and biologically meaningful information across species.

\begin{table}[!htb]
\centering
\caption{Dataset descriptions and statistics. $L$ denotes the total length [minutes], $S$ the number of samples, $n_\text{task}$ the number of classes, SR the sampling rate [kHz], $\mu$ the median length [ms].}
\begin{tabular}{llrrrrrrrr}
\toprule
\textbf{Dataset} & \textbf{Animal} & $\textbf{S}$ & $\bm{L}$ & \textbf{SR} & $\bm{n_\text{CTID}}$ & $\bm{n_\text{CLID}}$ & $\bm{n_\text{SID}}$ & $\bm{\mu}$ & $\bm{\sigma}$ \\
\midrule
IMV & Marmosets & $72,920$ & $464$ & $44.1$ & $11$ & $10$ & -- & $127$ & $375$ \\
Bosshard & Marmosets & $13,808$ & $37$ & $300$ & 7 & 8 & 2 & $117$ & $181$ \\
Wierucka & Marmosets & $4,901$ & $138$ & $125$ & 12 & 8 & 2 & $1,037$ & $1,687$ \\
Abzaliev & Dogs & $8,034$ & $137$ & $48$ & $14$ & $80$ & $2$ & $655$ & $1313$ \\
\bottomrule
\end{tabular}
\label{table:dataset_stats}
\end{table}

\textbf{InfantMarmosetsVox (IMV)} \citep{sarkar23_interspeech} is an extended version of the dataset used in the study on marmoset call type discrimination by ~\citep{cas_data}. The dataset consists of 72,920 audio segments representing 11 different call-types, and amounting to 464 minutes of vocalizations. The data contains 350 files of precisely labeled 10-minute audio recordings across all ten caller classes. The audio was recorded from five pairs of infant marmoset twins, each recorded individually in sound-proofed rooms at 44.1 kHz SR, without communication with other marmoset pairs or the experimenters. The audio recordings were manually labeled by an experienced researcher using the `Praat' tool. For each vocalization, the start and end time, call type, and marmoset identity are been provided. Although a large dataset by bioacoustics standards, each segment is predominantly short, at a median length of 127 ms. The spectral range of the calls is mostly centered at around 7-8 kHz, although there is still some information present above 16 kHz \citep{sarkar24_vihar}. The calltypes are entitled peep (pre-phee), phee, twitter, trill, trillphee, tsik tse, egg, pheecry (cry), trllTwitter, pheetwitter, and peep calls.
    
The \textbf{Bosshard} \citep{Alex_MA_thesis, bosshard_2024} dataset consists of 102 labeled 10-min focal audio recordings of common marmoset calls recorded in six behavioural contexts. A pair of marmosets was either separated or in the same enclosure, with preferred food either freely available for the focal individual or not. Each of the 8 subjects was recorded on 16 separate occasions. Most of the calls were given in bouts as holistic single call units, and thus, a call-type unit was defined as a call bout with call elements which were not further apart than 0.5s, as per existing literature \citep{Agamaite2015, Snowdon2001}. We only used the segments labeled as single call elements, i.e. not split up in bouts, to avoid data overlap and duplication. The dataset consists of 7 calls, namely alarm, ek, food, phee, trill, tsk, and twitter. The audio recordings were manually annotated by using Avisoft SASLab Pro (Avisoft Bioacoustics, Feb. 2017) to narrowly label the start and end of each call-type. The data was collected under Swiss legislation and licensed by Zurich's cantonal veterinary office (license ZH 223/16 and ZH 232/19).
    
The \textbf{Wierucka} dataset was collected from 6 target adult common marmosets, 3 male and 3 female, housed at the University of Zurich. Two additional non-target individuals were also included in the dataset, summing to 8 individuals in total. The data consists of 12 calls classes: phee, trill, food call, tsk, low tsk (tsk with a peak frequency of approximately 7-9 kHz), twitter (sequence), ek, phee sequence (multiple phees), low tsk sequence (multiple low tsks), ek sequence (multiple eks), food call sequence (multiple food calls). All procedures were done in accordance with Swiss legislation and were licensed by Zurich’s cantonal veterinary office (license ZH223/19). For each recording, two individuals (one male and one female) were placed in adjacent wire cages and recorded simultaneously in 15-minute intervals with two UltraSoundGate 116H recorders coupled with an Avisoft CM16/CMPA condenser microphone (Avisoft Bioacoustics, Germany), each set to a different gain to capture both low and high amplitude calls with a sampling rate of 125kHz. A total of 12 recordings, spread over 7 months, were made for each target individual. Caller identity was labeled in real time using Avisoft-RECORDER USGH (Avisoft Bioacoustics, Germany). The labelling of the calls’ exact start and end points was carried out through a visual examination of the spectrograms. For inclusion in subsequent analyses, calls needed be distinctly visible on the spectrogram, devoid of any interference from other calls, and readily classifiable into specific call-type categories.

Dog vocalizations offer another intriguing domain for bioacoustic research, where subtle differences in bark types and other sounds can convey distinct emotional states or intentions. In our study, we focus on datasets that capture a range of canine vocal behaviors—from aggressive or fearful barks to those associated with excitement or owner interaction. \textbf{Abzaliev} dataset is novel dog dataset (here referred to by the first author's name) consisting of 8,034 total vocalizations \citep{abzaliev24}. It contains 14 different call-types, ranging from normal, aggressive, fearful, and playful barks at strangers (IDs 0--3), to vocalizations related to owner interaction (4--5) and non-stranger/non-play sounds (6). It also contains postive or negative whines (7--8) and growls (9--10), barks associated with sadness or anxiety (11), and excitement upon the owner's arrival home (12). The recordings originate from various dog breeds, including Chihuahuas, French Poodles, and Schnauzers. The data was recorded at 48 kHz SR from a microphone, and followed a protocol designed and validated by experts in animal behavior. The dog vocalizations were induced by exposing the dogs to different types of external stimuli, with the participation of the owner and/or experimenter. We discard all the segments labeled as non-dog sounds, such as TV, cars, and appliances.

\section{Levenshtein Distance}

We use the Levenshtein distance $d(\bm{t}_1, \bm{t}_2)$, a string metric also known as the edit distance, to quantitatively measure the distance between a pair of discrete token sequences $\bm{t}_1$ and $\bm{t}_2$. The distance effectively represents the minimum number of `edits', i.e. insertions, deletions, or substitutions, needed to change one sequence into the other. A distance $d = 0$ thus means that the two sequences are identical. It can go up to at most the length of the longer string. However, this metric gives an absolute difference between sequences and is misrepresentative when a pair of sequences have a large difference in lengths. To overcome this issue, we use the normalized Levenshtein distance, which divides the calculated distance by the length of the longer sequence $\frac{d(\bm{t}_1, \bm{t}_2)}{\max(|\bm{t}_1|, |\bm{t}_2|)}$, where $|\cdot|$ denotes the length of the sequence. In this case, the distance is bounded between 0 and 1, representing identical and completely different sequences respectively. In the case of $d=1$, one need to edit every character in the longer string to transform it into the other.

\section{Classifier Experimental Setup}
\label{sec:classif_exp_setup}
We train the $k$-NN classifier by iterating over the hyperparameters given in \Cref{tab:knn_search_space} for each layer to obtain optimal classification results. The classifier is trained on \textit{Train}, and the hyperparameters defined in the search space are evaluted on \textit{Val}, using UAR as the optimization criterion. The best hyperparameters are used on \textit{Test}, and the predicted label of a sample is determined by applying a majority-voting framework on the actual labels of the $k$ most similar sequences.

\begin{table}[htb]
    \centering
    \caption{Hyperparameter search space used for training the $k$‑NN classifier.}
    \begin{tabular}{lll}
    \toprule
    \textbf{Classifier} & \textbf{Hyperparameter} & \textbf{Search Space} \\
    \midrule
    \multirow{4}{*}{$k$‑NN}
    & Number of neighbours $k$ & [1, 3, 5, 7, 9] \\
    & Neighbour weighting & [Uniform, distance] \\
    & Distance & Levenshtein \\
    & Task & [CTID, CLID] \\
    \bottomrule
    \end{tabular}
    \label{tab:knn_search_space}
\end{table}

\newpage
\section*{NeurIPS Paper Checklist}

\begin{enumerate}

\item {\bf Claims}
    \item[] Question: Do the main claims made in the abstract and introduction accurately reflect the paper's contributions and scope?
    \item[] Answer:  \answerYes{} 
    \item[] Justification: All claims in the abstract and introduction accurately reflect the paper's contributions and scope.

\item {\bf Limitations}
    \item[] Question: Does the paper discuss the limitations of the work performed by the authors?
    \item[] Answer:  \answerYes{} 
    \item[] Justification: Yes, the paper clearly shows the limitations of the proposed approach, but discusses potential for improvements.

\item {\bf Theory assumptions and proofs}
    \item[] Question: For each theoretical result, does the paper provide the full set of assumptions and a complete (and correct) proof?
    \item[] Answer: \answerNA{} 
    \item[] Justification: All results are justified, but there are no proofs needed or applicable to this applied work. 

    \item {\bf Experimental result reproducibility}
    \item[] Question: Does the paper fully disclose all the information needed to reproduce the main experimental results of the paper to the extent that it affects the main claims and/or conclusions of the paper (regardless of whether the code and data are provided or not)?
    \item[] Answer:  \answerYes{} 
    \item[] Justification: The paper includes detailed experimental setup, publicly available datasets, and in-depth description of proposed approaches.

\item {\bf Open access to data and code}
    \item[] Question: Does the paper provide open access to the data and code, with sufficient instructions to faithfully reproduce the main experimental results, as described in supplemental material?
    \item[] Answer:  \answerNo{} 
    \item[] Justification: The paper uses existing open-source datasets, as well as those available upon request. It also includes detailed description on the process to reproduce results. However, it does not include any source code.

\item {\bf Experimental setting/details}
    \item[] Question: Does the paper specify all the training and test details (e.g., data splits, hyperparameters, how they were chosen, type of optimizer, etc.) necessary to understand the results?
    \item[] Answer: \answerYes{} 
    \item[] Justification: All experimental details are given in the main text or in the appendix.

\item {\bf Experiment statistical significance}
    \item[] Question: Does the paper report error bars suitably and correctly defined or other appropriate information about the statistical significance of the experiments?
    \item[] Answer: \answerNo{} 
    \item[] Justification: The paper does not give error bars, but has clear training/val/test splits, as well as details on other optimization parameters. The experiments, however, were not running multiple times with an average and error bars for each result.

\item {\bf Experiments compute resources}
    \item[] Question: For each experiment, does the paper provide sufficient information on the computer resources (type of compute workers, memory, time of execution) needed to reproduce the experiments?
    \item[] Answer: \answerNo{} 
    \item[] Justification: The compute resources are relatively minor for a machine learning paper. They just involve feature extraction and classification. These do not require strong CPU/GPU resources, especially since bioacoustics datasets are small.
    
\item {\bf Code of ethics}
    \item[] Question: Does the research conducted in the paper conform, in every respect, with the NeurIPS Code of Ethics \url{https://neurips.cc/public/EthicsGuidelines}?
    \item[] Answer: \answerYes{} 
    \item[] Justification: The researech conducted in this paper fully respects the NeurIPS Code of Ethics.

\item {\bf Broader impacts}
    \item[] Question: Does the paper discuss both potential positive societal impacts and negative societal impacts of the work performed?
    \item[] Answer: \answerYes{} 
    \item[] Justification: The paper discusses the impact of this work for the bioacoustics community.
    
\item {\bf Safeguards}
    \item[] Question: Does the paper describe safeguards that have been put in place for responsible release of data or models that have a high risk for misuse (e.g., pretrained language models, image generators, or scraped datasets)?
    \item[] Answer: \answerNA{} 
    \item[] Justification: There is no real risk of misuse of the work conducted in this paper.

\item {\bf Licenses for existing assets}
    \item[] Question: Are the creators or original owners of assets (e.g., code, data, models), used in the paper, properly credited and are the license and terms of use explicitly mentioned and properly respected?
    \item[] Answer: \answerYes{} 
    \item[] Justification: The authors are original owners of one dataset, and the others are appropriately credited/cited. No other real assets were required or used in this work.

\item {\bf New assets}
    \item[] Question: Are new assets introduced in the paper well documented and is the documentation provided alongside the assets?
    \item[] Answer: \answerNo{} 
    \item[] Justification: This paper does not release any new assets.

\item {\bf Crowdsourcing and research with human subjects}
    \item[] Question: For crowdsourcing experiments and research with human subjects, does the paper include the full text of instructions given to participants and screenshots, if applicable, as well as details about compensation (if any)? 
    \item[] Answer: \answerNA{} 
    \item[] Justification: This paper does not involve crowdsourcing nor research with human subjects.

\item {\bf Institutional review board (IRB) approvals or equivalent for research with human subjects}
    \item[] Question: Does the paper describe potential risks incurred by study participants, whether such risks were disclosed to the subjects, and whether Institutional Review Board (IRB) approvals (or an equivalent approval/review based on the requirements of your country or institution) were obtained?
    \item[] Answer:  \answerNA{} 
    \item[] Justification: This paper does not involve crowdsourcing nor research with human subjects.

\item {\bf Declaration of LLM usage}
    \item[] Question: Does the paper describe the usage of LLMs if it is an important, original, or non-standard component of the core methods in this research? Note that if the LLM is used only for writing, editing, or formatting purposes and does not impact the core methodology, scientific rigorousness, or originality of the research, declaration is not required.
    \item[] Answer: \answerNA{} 
    \item[] Justification: The core method development in this research did not involve any LLMs as any important, original, or non-standard components. The research and proposed methods fully originate from the authors.

\end{enumerate}

\end{document}